\documentclass[runningheads]{llncs}
\usepackage[T1]{fontenc}
\usepackage{graphicx}
\usepackage{amssymb}
\usepackage{multirow}

\begin{document}
\title{Multimodal Machine Learning for Real Estate Appraisal: A Comprehensive Survey}
\titlerunning{Multimodal ML for  Real Estate Appraisal: A Survey}
\author{Anonymous Author}

%
\author{Chenya Huang\orcidID{0009-0008-6686-4393} \and
Bin Liang*\orcidID{0000-0002-6605-2167} \and
Zhidong Li\orcidID{0000-0003-3288-5547} \and
Fang Chen\orcidID{0000-0003-4971-8729}}
\authorrunning{C. Huang et al.}
\institute{University of Technology Sydney, Sydney, Australia\\
\email{chenya.huang@student.uts.edu.au} \\
\email{\{bin.liang, zhidong.li, fang.chen\}@uts.edu.au}
}

\maketitle              % typeset the header of the contribution
\begin{abstract}
Real estate appraisal has undergone a significant transition from manual to automated valuation and is entering a new phase of evolution. Leveraging comprehensive attention to various data sources, a novel approach to automated valuation, multimodal machine learning,  has taken shape.  This approach integrates multimodal data to deeply explore the diverse factors influencing housing prices. Furthermore, multimodal machine learning significantly outperforms single-modality or fewer-modality approaches in terms of prediction accuracy, with enhanced interpretability. However, systematic and comprehensive survey work on the application in the real estate domain is still lacking. In this survey, we aim to bridge this gap by reviewing the research efforts. We begin by reviewing the background of real estate appraisal and propose two research questions from the perspecve of performance and fusion aimed at improving the accuracy of appraisal results. Subsequently, we explain the concept of multimodal machine learning and provide a comprehensive classification and definition of modalities used in real estate appraisal for the first time. To ensure clarity, we explore works related to data and techniques, along with their evaluation methods, under the framework of these two research questions. Furthermore, specific application domains are summarized. Finally, we present insights into future research directions including multimodal complementarity, technology and modality contribution. 

\end{abstract}
\section{Introduction}

The prediction of housing prices has been an area of significant interest for over a century, evolving alongside advancements in economic theory, computational tools, and data availability. This field plays a crucial role in property sales, guiding investments, and informing policy decisions. Early approaches to real estate appraisal were predominantly manual, relying on subjective evaluations by appraisers and comparative analysis of similar properties. These methods, collectively referred to as empirical appraisal approaches \cite{zhang2021mugrep}. Thay are grounded in market dynamics and the three key elements of sales: cost, price, and profit. The most common methodologies included the sales comparison method, the income capitalization method, and the cost method.

While these empirical methods laid the foundation for property price prediction, they were constrained by several limitations, including high costs, lengthy evaluation processes, and susceptibility to human bias. Efforts to incorporate concepts such as the time value of money brought incremental improvements but failed to address these challenges comprehensively. To overcome these issues, Automated Valuation Models (AVMs) were introduced, leveraging quantitative techniques to estimate housing prices based on variables such as property size and economic indicators. This marked a pivotal shift towards data-driven approaches in housing price prediction.

\begin{figure}
    \centering
    \includegraphics[width=1\linewidth]{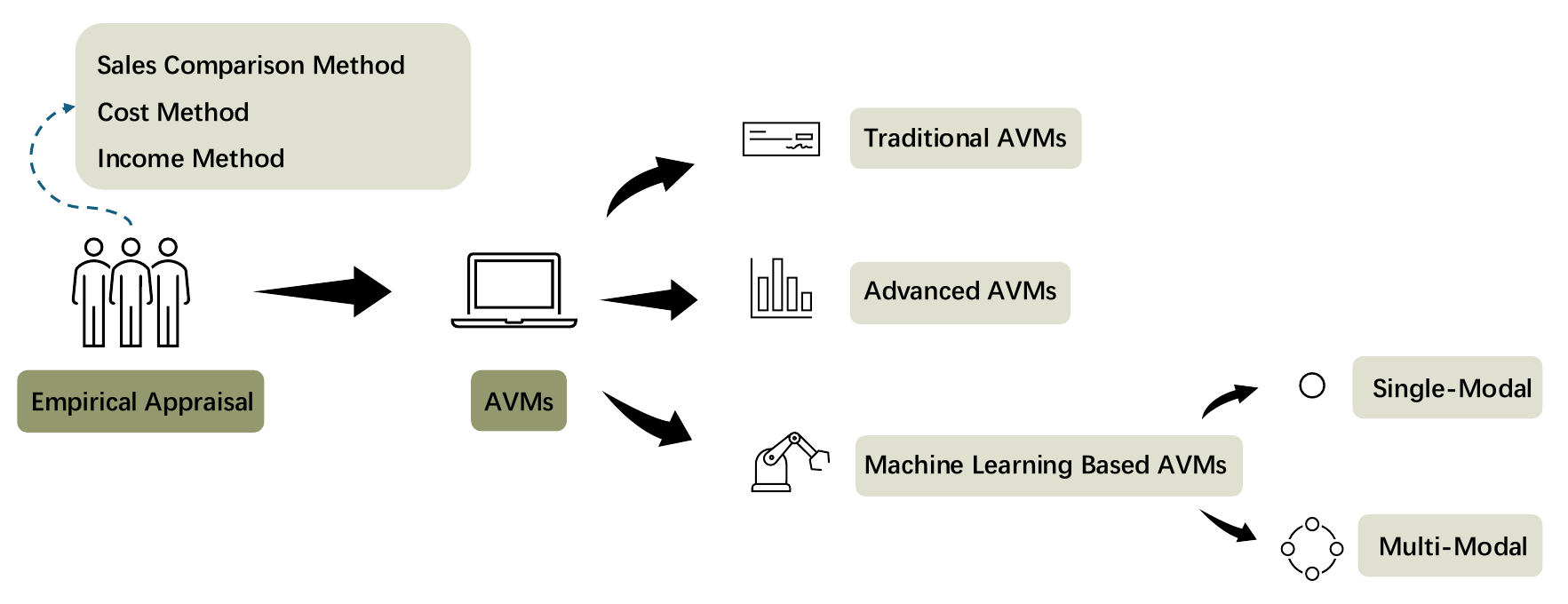}
    \caption{Development Trends in Real Estate Appraisal}
    \label{fig:develop}
\end{figure}

With the rise of machine learning, new hybrid models combining different approaches demonstrated significant improvements in predictive accuracy compared to traditional statistical models. Models such as ANN-GIS \cite{garcia2008ann+}, PSO-SVM \cite{wang2014real}, and advanced techniques have exceeded the capabilities of early neural networks. However, issues including overfitting, instability, and low interpretability remain challenges. Recently, the adoption of big data technologies has propelled house price prediction forward. By integrating diverse data types—including text, images, and geographic information—multimodal machine learning enables a comprehensive analysis of housing markets, capturing spatial and temporal dynamics. This kind of approach has improved both interpretability and predictive accuracy, establishing a new milestone in the field. These trends mentioned above are illustrated in Fig. \ref{fig:develop}, which summarizes the two major milestone developments in real estate valuation as well as the ongoing subfield of multimodality.

While existing articles review the progress in real estate appraisal, most are limited to early-stage machine learning methods and overlook recent advancements. This survey aims to bridge this gap by providing a comprehensive review of multimodal machine learning for this field. We begin by introducing the background of multimodal learning and the types of multimodal data commonly used in this domain. Next, we present a detailed overview of the modeling process, emphasizing two key research questions (RQ) for enhancing predictive accuracy: \textbf{RQ 1) model performance; RQ 2) modality fusion}. Subsequently, we categorize existing methods aligning with these two directions. Finally, we highlight current challenges in multimodal learning in real estate appraisal and outline potential future directions for advancing the field.

\section{Background on Multimodal Machine Learning}

A modality refers to the way humans perceive or experience events \cite{ngiam2011multimodal}, encompassing sensory categories like auditory, olfactory, and tactile, as well as forms of data acquisition, such as images, text, and radar \cite{liang2024foundations}. However, modalities present challenges due to their heterogeneity—distinct structures and representations—and their interdependence, requiring effective integration. Researchers have identified five core challenges in multimodal learning: representation, translation, alignment, fusion, and co-learning \cite{baltruvsaitis2018multimodal}. Building on this, Liang et al. \cite{liang2024foundations} introduced a new principle for modalities—their interaction in task reasoning can generate new information—and proposed a innovative classification way. In this updated framework, apart from representation and alignment, the other categories include reasoning, generation, transference, and quantification. To address these multimodal challenges,  researchers continue to explore and advance solutions in practical applications.

The concept of multimodality was first applied in the 1980s, primarily focusing on semantic-level integration. One prominent example is Bolt human-computer interaction system, ``Put-That-There'', which combined speech and gesture in a graphical interface \cite{bolt1980put}. It can be viewed as an early framework for multimodal audio-visual speech recognition.  Subsequently, a neural network-based multimodal fusion model demonstrated the feasibility of multimodal fusion by enhancing speech recognition with visual cues \cite{yuhas1989integration}. Frameworks by Ngiam et al. \cite{ngiam2011multimodal} enabled shared feature learning across modalities, extending applications beyond audio-visual tasks. Recent approaches combine architectures like LSTMs, RNNs, and CNNs to effectively merge temporal and spatial features for enhanced performance \cite{feng2017audio}.

The other applications include multimedia retrieval and group interaction analysis. Modern models like Two-Tower and CLIP \cite{radford2021learning} leverage shared representations for text and images, enabling deeper semantic understanding. Retrieval-augmented models now enhance task performance by improving reasoning and answer generation. The current research trend is focused on leveraging large models to enhance the application of multimodal learning in retrieval. In group interaction analysis,  the AMI Meeting Corpus \cite{carletta2005ami} have been provided which includes 100 hours of multimodal data, such as recordings, videos, and whiteboard notes. It used to monitor the dominant participants in the meeting. Recently, sensor data has also been incorporated to analyze social signals for understanding team actions \cite{lehmann2024multimodal}. These research efforts have led to significant advances in emotion analysis and non-verbal emotional feedback.

Multimodal interaction has evolved into modern intelligent agents, transitioning from simple bimodal combination models to multimodal fusion with Transformer-based large models. Emerging trends are flourishing in the application domains of multimodal learning, featuring a greater diversity of modalities and larger dataset scales. MultiBench \cite{liang2021multibench} now provides comprehensive benchmark datasets across 15 domains and 10 modalities, including novel elements such as force sensors and proprioception sensors. This progression has gradually addressed prior challenges.Additionally, the range of application scenarios is expanding, transitioning from theoretical frameworks to practical industries such as healthcare, finance, real estate, and manufacturing.

\section{Multimodal Machine Learning in Real Estate Appraisal}
\begin{figure}
    \centering
    \includegraphics[width=1\linewidth]{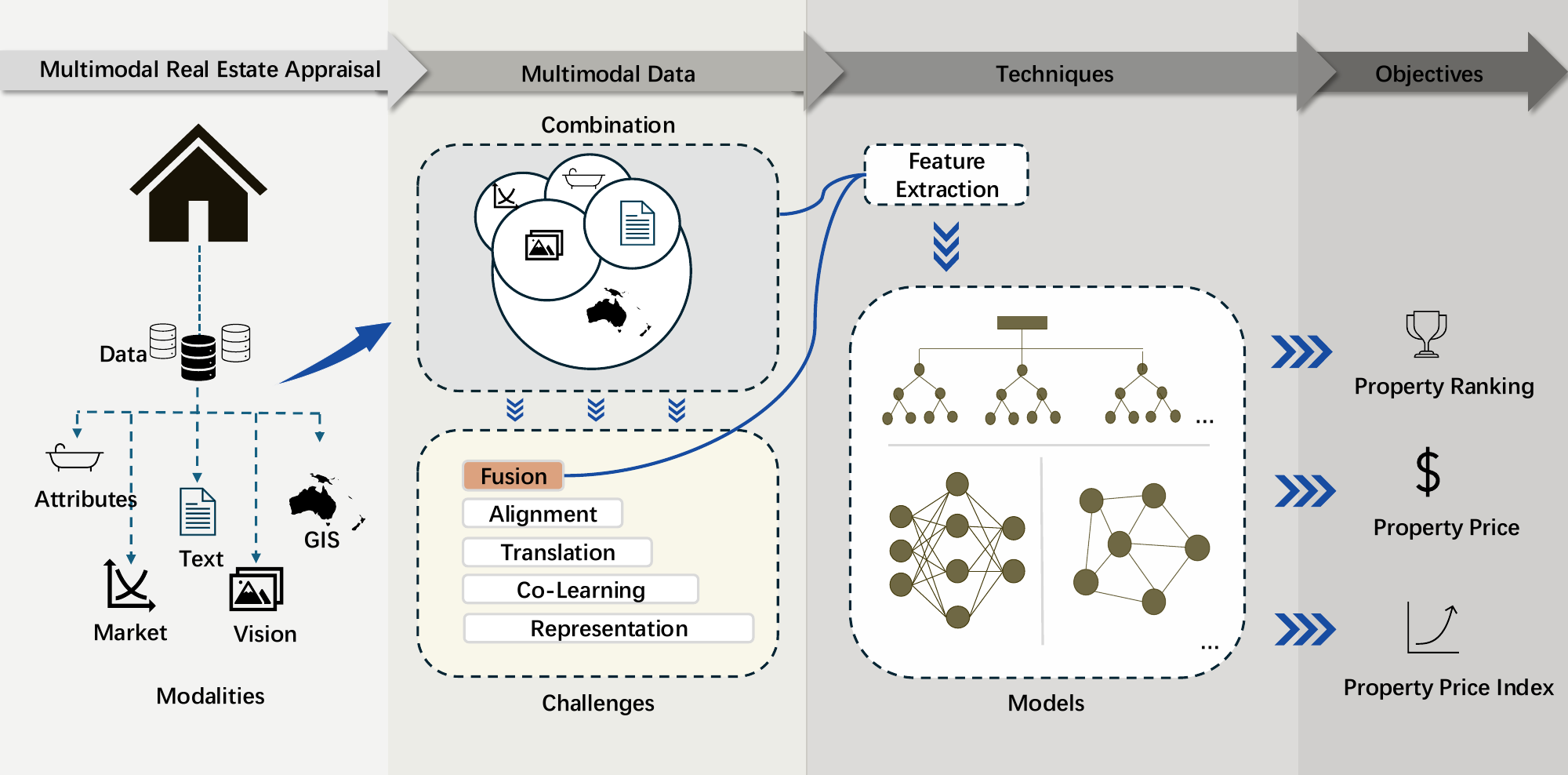}
    \caption{Overview Framework for the Survey}
    \label{fig:framework}
\end{figure}

This section begins by introducing the multimodal data involved in housing price prediction. Following this, we review the multimodal techniques and models referenced in this field and outline the methods for evaluating performance and accuracy. These components are presented in framework Fig.\ref{fig:framework} and can be categorized into four sections, encompassing the entire workflow from modality collection, processing, modeling, to output generation.

\subsection{Multimodal Data }

In current real estate appraisal research, five common modalities of property-related data are primarily used for predicting housing prices: attributes data, market data, textual data, visual data and Geographic Information System (GIS) data. Heterogeneous tabular data is one of the most common data types. Before other modal data became collectible, tabular data dominated research datasets \cite{borisov2022deep}. Even today, tabular data remains an important component of real estate appraisal. In early house price appraisals, tabular data primarily consisted of attribute data and market data. Based on their sources and corresponding subject objects, we formally categorize them into two distinct modalities for further discussion.

\textbf{Attributes data} refers to the characteristics of a property, which includes both continuous and categorical data. Examples include the number of bedrooms, area, location, floor level, presence of a swimming pool, and availability of parking \cite{zhang2023reinforcement} \cite{garcia2008ann+}. However, there are exceptions in the data collection process, such as the detailed address of the property, which is a text string. Although inherently textual data, it does not fall under the modality we will discuss next. Further explanation will follow.

\textbf{Market data} refers to information sourced from the real estate market, and occasionally from financial and economic markets. This includes, but is not limited to, historical property sales data (transaction volume and price trends), interest rates (deposit and loan rates)\cite{dupre2020urban}, property tax\cite{de2018economic} and GDP growth rates\cite{ge2019lstm}.  In addition, rent-related data has also been taken into consideration. Potrawa et al. \cite{potrawa2022much} suggests that the level of rent could influence future housing prices. These data are typically in the form of time series.

\textbf{Visual data} conveys the basic information of a property, often in a more intuitive and memorable way. It referred to here specifically includes image data, without yet encompassing video data. Some image-based studies highlight different perspectives. In these studies, visual features include photos of bedrooms, kitchens, bathrooms, and the house frontal view \cite{yousif2023real}\cite{poursaeed2018vision}\cite{nouriani2022vision}. Additionally, other researches use street and window views to assess property values\cite{law2019take}. Most existing research on image processing can be categorized into two approaches: classification (or rating) based on specific features, and feature extraction for vector transformation. Poursaeed et al.\cite{poursaeed2018vision} used crowdsourcing to estimate luxury levels based on room types and trained DenseNet to categorize real estate images into eight luxury levels. In contrast, Kang et al. \cite{kang2019effects} focused on the broader external environment of properties, calculating a livability score for each neighborhood and assigning a perceived score from 1 to 10. To optimize feature extraction, more refined categorizations have been introduced, such as color classification and interior-exterior differentiation \cite{kostic2020image}\cite{law2019take}\cite{potrawa2022much}. However, both methods have limitations. While rating classification is easily comprehensible, it is time-consuming. Feature extraction, though more efficient, lacks interpretability and is challenging to monitor, with performance assessment relying solely on output evaluation. While research in image processing continues to advance, the field primarily relies on these two established approaches. Innovative methods are yet to gain widespread adoption in the domain of real estate image analysis.

\textbf{Textual data} is also included. Pryce \cite{pryce2013rhetoric} showed that the content and rhetoric of promotional messages in advertisements significantly influence marketing outcomes and sales prices. Similarly, sentiment texts such as news, blogs, reviews, and comments also have economic impacts on property prices. Kou \cite{kou2018understanding} explored the relationship between media sentiment indices and housing price indices at a micro level using newspaper articles and internet data. Analogously, Zhao et al. \cite{zhao2022pate} uncovered more detailed patterns of influence. The third type of textual data is the most common. It provides a direct description of various house characteristics, without involving subjective emotions or marketing exaggeration. Examples include detailed features of houses corresponding to attribute data  \cite{zhang2024describe}\cite{yousif2023real}\cite{baur2023automated}. Gentzkow et al. \cite{gentzkow2019text} provide a comprehensive overview of methodologies for transforming textual data into analyzable formats, highlighting common approaches such as the Bag-of-Words \cite{shahbazi2016estimation} and TF-IDF\cite{zhang2024describe}. These transformations serve as a foundation for advanced analytical techniques, including text regression models like Lasso Regression \cite{baur2023automated}, and word embedding approaches such as Word2Vec \cite{shahbazi2016estimation}, and BERT \cite{baur2023automated}. 

\textbf{GIS data} has gradually been incorporated into research in recent years to better capture the relationship between housing prices and spatial factors. Fu et al. \cite{fu2016modeling}suggested that geographical dependence is the key to housing appraisal. Initially, GIS data accounted for only a small portion of attributes or textual data, often appearing as a detailed address, a regional name, or simply as latitude and longitude\cite{baur2023automated}. Over time, additional elements have been incorporated into this data category. As of now, transportation data \cite{antipov2012mass}\cite{tan2016modeling}, population\cite{ge2019integrated}, points of interest (POI) \cite{fu2016modeling}, and even remote sensing imagery \cite{ge2019integrated} collectively form the standard configuration of a GIS dataset. The most common methods involve calculating the distances between properties and POI or transportation, or simply counting the number of such facilities to evaluate convenience in relation to price. Moreover, studies have utilized population mobility distribution maps to estimate the popularity of residential developments \cite{fu2016modeling}\cite{zhang2021mugrep}. 

From Table 1, it is evident that since 2014, market, textual, and visual data have gradually been recognized as commonly used modalities closely associated with housing prices. This trend is linked to advancements in technology, as deep learning capable of efficiently processing such data matured during that period. With technological support, it is a direction worth exploring to implement multimodal approaches in housing price research.
\begin{table}[ht]
\caption{The distribution of modalities in real estate appraisal research}
\centering
\label{t:rs_source}
\begin{tabular}{ccccccc}
\hline
 Publication & Year & Attributes\footnotemark[1] & Market\footnotemark[2] & Text\footnotemark[3] & Vision\footnotemark[4] & GIS\footnotemark[5] \\ \hline
 García et al.\cite{garcia2008ann+}  & 2008 & \checkmark &  &  &  & \checkmark \protect \\
 Fu et al. \cite{fu2014sparse} & 2014 & \checkmark & & \checkmark & & \checkmark \protect \\
 Tan et al. \cite{tan2016modeling} & 2016 & \checkmark & & \checkmark & & \checkmark \protect \\
 De et al.\cite{de2018economic}  & 2018 & \checkmark & \checkmark & \checkmark  & \checkmark & \checkmark \protect \\
 Poursaeed et al. \cite{poursaeed2018vision} & 2018 & \checkmark & & & \checkmark & \protect \\
 Ge et al.\cite{ge2019lstm}  & 2019 & \checkmark & \checkmark &  &  & \checkmark \protect \\
 Law et al.\cite{law2019take} & 2019 & \checkmark & & & \checkmark & \checkmark \protect \\
 Dupré et al.\cite{dupre2020urban}  & 2020 & \checkmark & \checkmark &  &  & \checkmark \protect \\
 Kostic et al.\cite{kostic2020image} & 2020 & \checkmark &  & \checkmark & \checkmark & \checkmark \protect \\
 Wang et al.\cite{wang2021joint} & 2021& \checkmark & \checkmark &  &  & \checkmark \protect\\
 Zhang et al.\cite{zhang2021mugrep} & 2021 & \checkmark & \checkmark &  &  & \checkmark \protect\\
 Nouriani et al.\cite{nouriani2022vision} & 2022 & & \checkmark & \checkmark & \protect \\
 Potrawa et al.\cite{potrawa2022much} & 2022 & \checkmark & \checkmark &  & \checkmark & \checkmark \protect \\
 Zhang et al.\cite{zhang2024describe} & 2024 & \checkmark & &\checkmark &  & \protect \\ \hline
\end{tabular}\\
\raggedright
\begin{footnotesize}
\noindent\footnotemark[1] Attributes refers to Characteristics of the property.\\
\noindent\footnotemark[2] Market refers to Market economic data .\\
\noindent\footnotemark[3] Text refers to Property descriptions, advertisements and reviews.\\
\noindent\footnotemark[4] Vision refers to Interior and exterior photos of the property and street view photos .
\noindent\footnotemark[5] GIS refers to Geographic data including POI, transportation, regional population and income.\\

\end{footnotesize}
\end{table}

\subsection{Techniques and Models}

\begin{figure}
    \centering
    \includegraphics[width=1\linewidth]{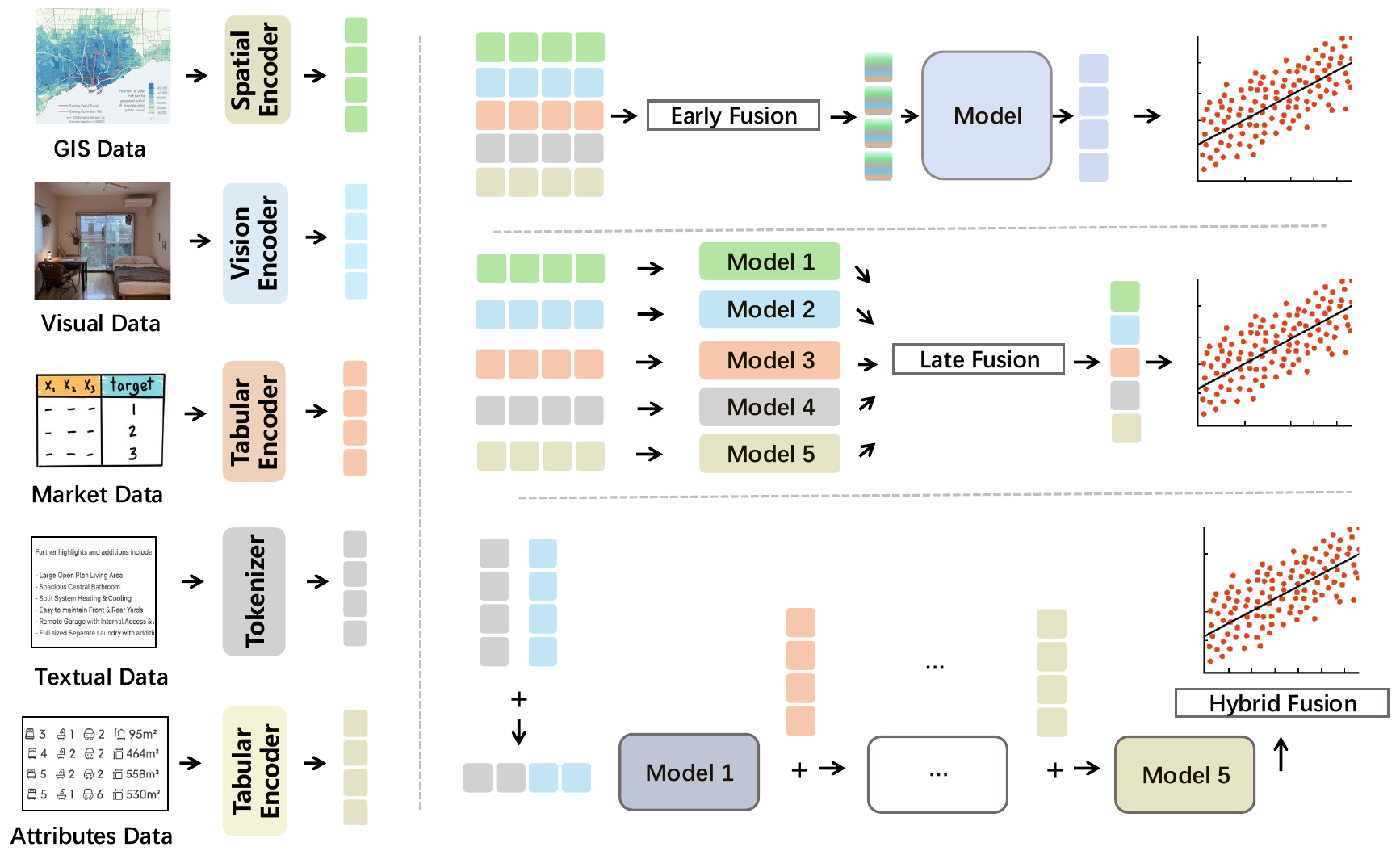}
    \caption{Fusion Approaches: Early Fusion, Late Fusion and Hybrid Fusion}
    \label{fig:fusion}
\end{figure}

Multimodal learning can be likened to a coin toss: one side represents challenges, while the other unveils core techniques as solutions. These two aspects are inherently intertwined. In the real estate sector, data modalities are both complex and diverse. This makes fusion one of the most critical challenges. Its corresponding techniques have garnered significant attention. Fusion can be broadly categorized into three types: early fusion, late fusion, and hybrid fusion, as illustrated in Fig.~\ref{fig:fusion}. These techniques aim to address RQ 2. Frameworks or models built on these theoretical foundations can resolve RQ 1. Together, they ultimately enhance overall prediction accuracy.
The primary difference between early fusion and late fusion lies in the timing of fusion \cite{gadzicki2020early}. Early fusion occurs at the input stage, where raw data or extracted features are combined into a unified representation before being fed into the model. In contrast, late fusion takes place at the output level, where the predictions from each modality are aggregated through weighted integration. Both approaches have their strengths and limitations. Early fusion excels at capturing deep interactions between modalities, whereas late fusion is better suited for tasks with strong modality independence. If the data quality of a certain modality is low, the performance of early fusion may suffer significantly. In contrast, late fusion remains robust. To address this, hybrid fusion combines the strengths of early and late fusion by integrating data from different modalities across multiple levels. In real estate price appraisal, all three methods have been utilized. 

The development of multimodal machine learning in the real estate sector has been relatively standard, following the basic progression of machine learning. Traditional machine learning models and deep learning models are currently being applied in combination. Based on the nonlinear relationship between housing prices and various multimodal factors, Shim et al. \cite{shim2014semiparametric} proposed the SSELS-SVM model, which is based on Support Vector Machines (SVM), to measure and predict house sale prices. Additionally, ensemble learning models such as XGBoost \cite{de2018economic}\cite{dupre2020urban}\cite{zhao2022pate}, Gradient Boosting \cite{baur2023automated}, and Random Forest \cite{antipov2012mass}\cite{rico2021machine} have also been widely used. Alkan et al.  \cite{alkan2023using} compared KNN, RF, and SVM, concluding that the SVM algorithm achieved the most successful results. However, most of these models face similar challenges, particularly in handling high-dimensional and complex data. For instance, in multimodal settings with heterogeneous data, fusion may lead to dimensionality issues, resulting in the so-called "curse of dimensionality." To overcome these challenges, neural networks have gained widespread favor in this area, particularly GNN and CNN. An automated real estate appraisal system combining GIS and ANN was proposed by García et al. \cite{garcia2008ann+}. Additionally, JGC\_MMN \cite{wang2021joint}, ST-RAP \cite{lee2023st}, and MugRep \cite{zhang2021mugrep} utilize CNN and GNN for feature extraction and fusion. However, the interpretability of deep learning is relatively low. To address this, some researchers have combined traditional methods with deep learning to improve prediction performance. For instance, Zhan et al.\cite{zhan2023hybrid} proposed a hybrid machine learning framework. Although multimodal machine learning has advanced into the era of large models, its application in the real estate sector has not kept pace with this progress. It has only just begun the initial exploration of Transformer framework\cite{moghimi2023rethinking}.

\subsection{Evaluation and Baseline}

In this survey, we address two types of research questions in house price prediction. For each type, the evaluation criteria for assessing the effectiveness of methods are divided into two corresponding groups. Before discussing these, it is essential to introduce commonly used metrics: $R^2$, Mean Absolute Error (MAE), and Root Mean Square Error (RMSE).

$R^2$ indicates the ability of a model to explain the variance of the target variable. A value closer to 1 demonstrates stronger explanatory power and better model fit. However, under the complex nonlinear relationships in multimodal settings, additional metrics are required to comprehensively evaluate model performance. For instance, Mugrep \cite{zhang2021mugrep} combines MAE and RMSE to evaluate overall performance. While these metrics can provide reasonable insights into the performance of a specific model, a true assessment requires horizontal comparisons. Based on existing research, horizontal comparisons can be categorized into two types: within-category comparisons and cross-category comparisons.

\paragraph{\textbf{Performance Review}}
A within-category comparison involves research models and baseline models belonging to the same category, for example, random forest and XGBoost are tree-based models. Hjort et al. \cite{hjort2024locally} proposed LitBoost and used GBT as the baseline model. Experimental results indicated that LitBoost outperformed GBT in overall experimental performance. However, it struggled to surpass GBT in practical applications. Similarly, Dupré \cite{dupre2020urban}conducted a more detailed comparison within the same category, focusing on boosters. In experiments using Dublin housing data, both tree-based XGBoost and linear regression-based XGBoost were applied. The results demonstrated that tree-based XGBoost exhibited significant advantages in handling multimodal features.  However, most studies emphasize cross-category comparisons. Compared to within-category comparisons, it provide a more comprehensive evaluation of whether a model outperforms all other types of models.

Cross-category comparisons are commonly used in most evaluations. For example, Antipov \cite{antipov2012mass} compared Random Forest with various types of models, such as multiple regression, MLP, and KNN. This comparison validated its superiority over most mass appraisal methods available in the market. Similarly, Random Forest was also compared with basic neural networks (Multilayer Perceptron) and boosting models \cite{rico2021machine}. The result revealed that Random Forest could not quantify the nonlinear relationship between house prices and features, prompting the recommendation of deep artificial neural network techniques. Methods like LGCN  \cite{ge2019lstm}and MugRep \cite{zhang2021mugrep}, which are GNN-based and incorporate LSTM or hierarchical heterogeneous community graph convolution modules respectively, validated this suggestion. Additionally, similar approaches such as ST-RAP \cite{lee2023st} used MugRep, ReGram, GCN, LR, and SVM as baseline models for comparison. 

\begin{table}[ht]
\centering
\caption{Comparison of Models Across Publications}
\begin{tabular}{ c c c c c } 
\hline
\multicolumn{1}{c}{\textbf{Comparison Type}} & 
\multicolumn{1}{c}{\textbf{Publication}} & 
\multicolumn{1}{c}{\textbf{Models}} & 
\multicolumn{1}{c}{\textbf{Result}} & 
\multicolumn{1}{c}{\textbf{Metrics}} \\ \hline

\multirow{5}{*}{Within-category} 
 & \multirow{2}{*}{\centering \cite{dupre2020urban}} & \textbf{Tree-based XGBoost} & \textbf{0.45} & \multirow{2}{*}{$R^2$}\\  
 &  & LR-based XGBoost & 0.21 & \\ \cline{2-5}
 & \multirow{3}{*}{\centering \cite{zhan2023hybrid}} & HBOS-XGBoost & 0.9467 & \multirow{3}{*}{$R^2$} \\  
 &  & \textbf{HBOS-CatBoost} & \textbf{0.9501} & \\ 
 &  & HBOS-AdaBoost & 0.9483 & \\ \hline

\multirow{20}{*}{Cross-category} 
 & \multirow{6}{*}{\centering \cite{antipov2012mass}} & \textbf{RF} & \textbf{14.86\%} & \multirow{6}{*}{MAPE \%} \\  
 &  & CHAID & 16.92\% & \\  
 &  & CART & 17.36\% & \\  
 &  & MLP & 20.53\% & \\  
 &  & KNN & 18.53\% & \\  
 &  & Regression & 18.33\% & \\ \cline{2-5}
 & \multirow{4}{*}{\centering \cite{ge2019integrated}} & VAR & 33.41 & \multirow{4}{*}{RMSE} \\  
 &  & SVR & 31.61 & \\  
 &  & ST-ANN & 30.59 & \\  
 &  & \textbf{D6-L9-C-f} & \textbf{22.81} & \\ \cline{2-5}
 & \multirow{5}{*}{\centering \cite{zhang2021mugrep}} & LR & 0.4776 & \multirow{5}{*}{MAE} \\ 
 &  & SVR & 0.4427 & \\  
 &  & GBRT & 0.364 & \\  
 &  & DNN & 0.355 & \\  
 &  & \textbf{MugRep} & \textbf{0.3244} & \\ \cline{2-5}
 & \multirow{5}{*}{\centering \cite{lee2023st}} & LR & 105.09 & \multirow{5}{*}{MAE} \\  
 &  & SVR & 90.31 & \\ 
 &  & Mug-Rep & 35.46 & \\ 
 &  & ReGram & 34.61 & \\ 
 &  & \textbf{St-RAP} & \textbf{21.77} & \\ \hline
\end{tabular}
\end{table}

\paragraph{\textbf{Review on Ablation Study}}
The mentioned quality specifically refers to the combination quality of multimodal data. To evaluate the quality of results derived from different combinations, an experimental method called ablation study is often used. Most studies typically use a single modality as a benchmark and incrementally add modalities to explore three key questions: 
\begin{enumerate}
    \item Whether multimodal approaches yield more accurate results?
    \item Which combinations of modalities perform better?
    \item How interactions between modalities influence the outcomes?
\end{enumerate}

\begin{figure}
\centering
\includegraphics[width=1\linewidth]{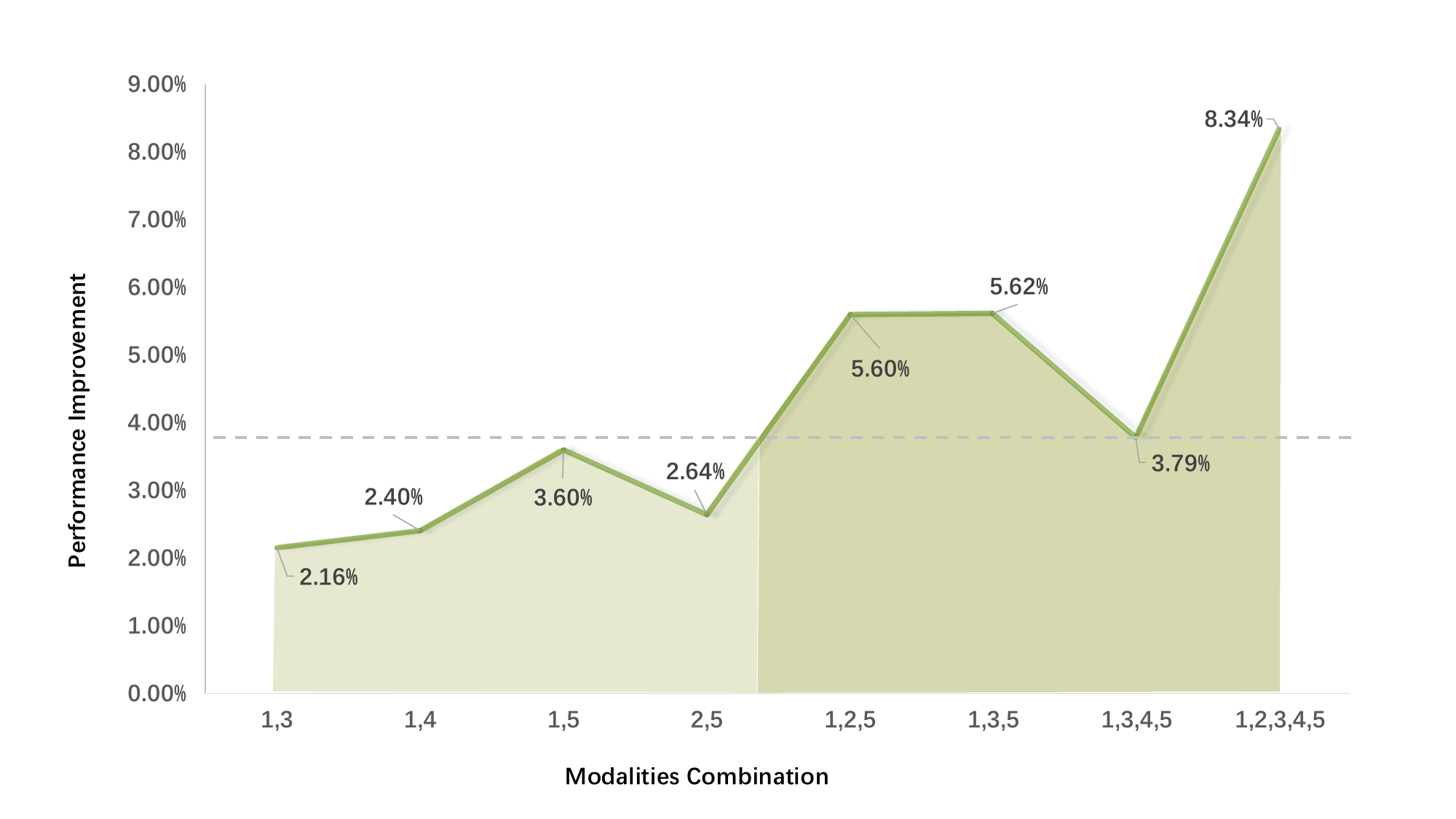}
\caption{Performance Improvement under Different Modalities Combinations.}
\label{fig:ablation_analysis}
\vspace{1em} 
\begin{minipage}{\linewidth}
\footnotesize
Note: 1 refers to attributes data, 2 refers to market data, 3 refers to textual data, 4 refers to visual data, 5 refers to GIS data.
\end{minipage}
\end{figure}

Zhang et al. \cite{zhang2024describe}treated textual features as a controlled variable and designed three combinations: textual features only, non-textual features only, and all features combined. When using only textual features, the $R^2$ reached 0.79. The model performed best when combined with attribute features. They concluded that textual and numerical features are complementary. Baur et al. \cite{baur2023automated}reached a similar conclusion even earlier. They further analyzed combinations of property description length with different price ranges. The results showed that textual descriptions contributed more significantly to predictions in the medium and high price segments. Moreover, textual descriptions were particularly effective in compensating for missing geographic information, such as latitude and longitude. We also observed that this phenomenon becomes more pronounced as the number of modalities increases. For example, PATE\cite{zhao2022pate} considered four features, including emotions, and tested five combinations. The experiments showed that adding any modality improved performance compared to using a single modality, with the best results achieved when all modalities were combined. However, we found that the transportation feature was overly simplistic, as it only included average traffic speed. This fails to adequately represent the primary relationship between real estate and geographic location. Other researches incorporated visual information into ablation studies\cite{poursaeed2018vision}\cite{yousif2023real}, yielding similar conclusions. Fig.~\ref{fig:ablation_analysis} illustrates how the results derived from all modalities in these studies outperform those obtained using a single modality in terms of accuracy. For example,  in \cite{zhang2024describe}, The combination of attributes and text results in an accuracy improvement of 2.16\% compared to using only attributes.  It is evident that the accuracy of tri-modal combination significantly surpasses that of bi-modal ones. When approaching full modality, the performance improvement reaches its peak. These studies demonstrated two unique enhancement modes in multimodal learning: supplementary enhancement and complementary enhancement\cite{jangra2023survey}. This indicates that the relationships between different modalities can  support and complement one another. 

\section{Applications}

Real estate appraisal has historically focused on single property appraisal. This is a highly personalized valuation service that requires assessors to determine the fair value of a property based on specific information and professional expertise\cite{zhang2021mugrep}. To address the issue of subjectivity in appraisal results, automated valuation models and machine learning have been introduced into real estate assessment. These advancements not only enhance productivity but  enable widespread application in mass appraisal. While serving different purposes, they complement each other to some extent. Single property appraisal focuses on market transactions at a micro level, while mass appraisal supports government taxation or market trend analysis from a macro perspective\cite{antipov2012mass}\cite{fan2019monopoly}. With the integration of multimodal learning, the two approaches are increasingly converging, achieving large-scale efficiency while enhancing appraisal accuracy through personalized algorithms. 

\begin{figure}
    \centering
    \includegraphics[width=0.75\linewidth]{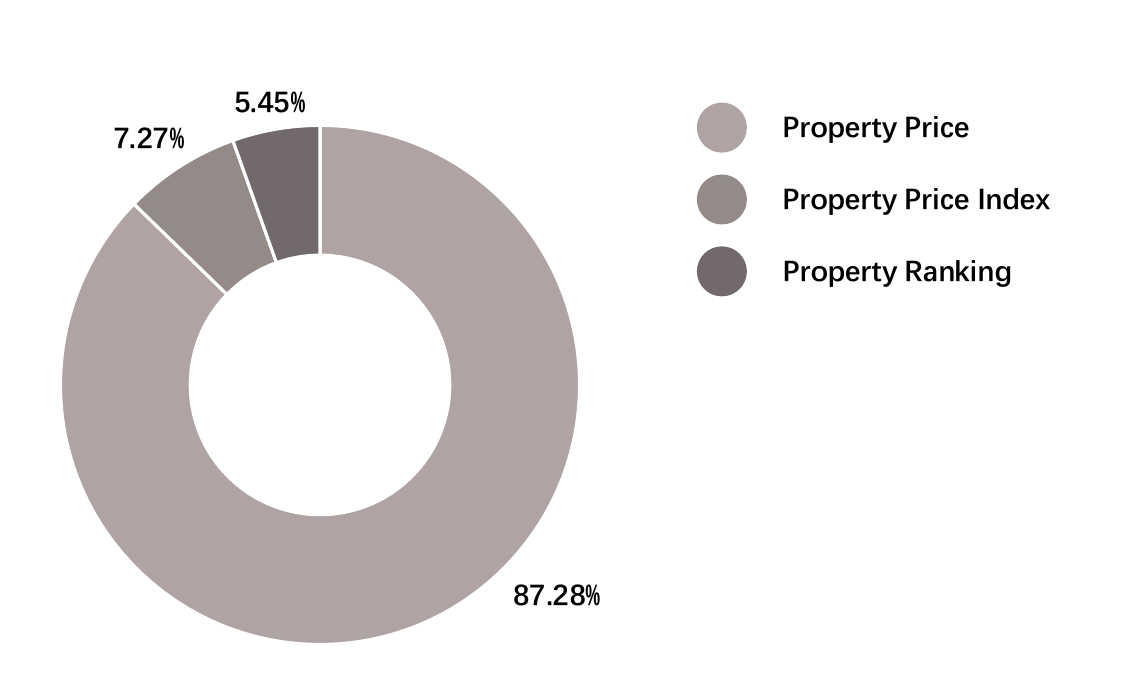}
    \caption{Main Applications Distribution}
    \label{fig:applications}
\end{figure}

However, in these appraisal studies, housing price is not the sole focus of appraisal. The Housing Price Index can also be used to track dynamic changes in housing prices. Several researchers \cite{kou2018understanding} have used multimodal features to predict the Residential Housing Price Index, analyzing fluctuations in residential housing prices and market trends. In contrast, Fu et al.\cite{fu2014sparse}\cite{fu2016modeling} focused on studying the ranking of real estate projects to assist different user groups in assessing investment value. In this process, they integrated online user reviews, land functionality, and geographic dependencies to construct a comprehensive property ranking model. In Fig.\ref{fig:applications}, we can see that the hot objective is predicting property price (over 85\%). This is because the supply-demand relationship remains the cornerstone of the real estate market, with vast consumers primarily focused on price fluctuations. This demonstrates that housing prices remain a highly trending topic with consistently high attention.

\section{Future Directions}

Although real estate appraisal using multimodal learning has made some progress, this field still faces unresolved challenges. In this section, we identify these challenges and discuss key research directions worth exploring in the future. Similarly, we approach this analysis from both data and technique perspectives. 

\subsubsection{Multimodal data enhancement potential }
Since the widespread study of data began, improving data quality has remained a persistent topic, particularly in the context of mixed multimodal data. In the previous discussion, we concluded that existing methods can independently process each modality and extract relatively reliable features. However, since the introduction of CCS-MMS \cite{jangra2021multi}, a new shift has emerged. The various modalities are not entirely independent but exhibit significant enhancement effects. This implies that, during the pre-processing stage, simply removing missing values is not sufficient. Missing information in one modality can be complemented by another, such as when room details are absent in attribute data but present in textual descriptions. However, no universal framework currently exists to capture and utilize these enhancement effects. Thus, there is substantial potential for exploration in this area. 

\subsubsection{Up-to-date Technology}
At present, most studies remain focused on generating predictions using neural networks, with some still investigating variations of the Hedonic Pricing Model. However, these techniques have already been extensively practiced and applied. Multimodal learning, on the other hand, has advanced into the era of large models. In real estate appraisal, the latest attempts with large model technologies have only reached the Transformer architecture, a deep learning model proposed back in 2017. 

\subsubsection{Evaluate the contribution of modalities}
Although ablation studies help identify the most effective modality combinations, it cannot accurately determine the specific contribution of an individual modality. Few studies have addressed this aspect. The purpose of ablation studies is to help select relevant modalities and optimize the model. The calculation of contribution levels is more detailed, considering all modalities to determine their specific impact on the prediction results. They complement each other on both a holistic and localized level. Moreover, their integration can enhance the interpretability of current deep learning models, improving the long-standing limitation of low explainability.

\section{Conclusion}

Real estate appraisal is a field with strong practical demands, whose enhanced scalability is attributed to the integration of multimodal machine learning. It has revolutionized previous approaches by capturing multiple factors that determine housing prices, moving beyond traditional linear relationships, which have been proven to be unrepresentative in housing price predictions. Given the lack of a comprehensive review on recent advancements in this subfield, this paper surveyed recent innovations in multimodal theories and technologies within real estate appraisal. The results of these innovations are categorized into two evaluation criteria: model performance and modality fusion. For each category, we introduced subcategories, such as dividing the currently involved modalities into five types, to provide a clearer understanding of the detailed developments in the multimodal real estate domain. However, we found that the application of multimodal technology in real estate has not fully realized its true potential at the forefront of innovation. For instance, there has been no detailed classification of modalities, and the models currently in use are relatively outdated. Hence, the second focus of this survey is to explore the current drawbacks in this field that urgently need improvement. Future research should prioritize the synergistic relationships between modalities, where one modality can interact with or complement another to some extent. This interaction is critical for achieving data fusion and alignment. In addition, determining the specific contribution of each modality to the prediction results requires detailed discussion. We believe that future studies should focus more on the implementation of cutting-edge multimodal technologies to address these evident challenges in this field.


\bibliographystyle{splncs04}
\bibliography{refs}
\end{document}